\documentclass[iicol]{sn-jnl}

\usepackage{graphicx}%
\usepackage{multirow}%
\usepackage{amsmath,amssymb,amsfonts}%
\usepackage{amsthm}%
\usepackage{mathrsfs}%
\usepackage[title]{appendix}%
\usepackage{xcolor}%
\usepackage{textcomp}%
\usepackage{manyfoot}%
\usepackage{booktabs}%
\usepackage{algorithm}%
\usepackage{algorithmicx}%
\usepackage{algpseudocode}%
\usepackage{listings}%

\usepackage[caption=false,font=normalsize,labelfont=sf,textfont=sf]{subfig}
\usepackage{stfloats}
\usepackage{url}
\usepackage{verbatim}
\usepackage{graphicx}
\usepackage{natbib}
\usepackage{booktabs}
\usepackage{wrapfig}
\usepackage{array}
\usepackage{tabularx}
\usepackage{multirow}
\usepackage{pifont}
\usepackage{xspace}
\usepackage{graphicx}
\newcommand{\cmark}{\ding{51}}%

\newcommand{\etal}{\textit{et al.}\@\xspace}
\newcommand*{\eg}{\textit{e.g.}\@\xspace}
\newcommand*{\ie}{\textit{i.e.}\@\xspace}

\def\tsc#1{\csdef{#1}{\textsc{\lowercase{#1}}\xspace}}
\tsc{WGM}
\tsc{QE}
\newcommand{\revise}[1]{\textcolor{black}{#1}}

\begin{document}

\title[Article Title]{Cross-Layer Attentive Feature Upsampling for Low-latency Semantic Segmentation}


\author[1]{\fnm{Tianheng} \sur{Cheng}}
\author[1]{\fnm{Xinggang} \sur{Wang}}
\author[1]{\fnm{Junchao} \sur{Liao}}
\author*[1]{\fnm{Wenyu} \sur{Liu}}\email{liuwy@hust.edu.cn}


\affil[1]{\orgdiv{School of Electronic Information and Communications}, \orgname{Huazhong University of Science and Technology}, \orgaddress{\city{Wuhan}, \postcode{430074}, \state{Hubei}, \country{China}}}




\abstract{Semantic segmentation is a fundamental problem in computer vision and it requires high-resolution feature maps for dense prediction.
Current coordinate-guided low-resolution feature interpolation methods, \eg, bilinear interpolation, produce coarse high-resolution features which suffer from feature misalignment and insufficient context information.
Moreover, enriching semantics to high-resolution features requires a high computation burden, so that it is challenging to meet the requirement of low-latency inference.
We propose a novel Guided Attentive Interpolation (GAI) method to adaptively interpolate fine-grained high-resolution features with semantic features to tackle these issues.
Guided Attentive Interpolation determines both spatial and semantic relations of pixels from features of different resolutions and then leverages these relations to interpolate high-resolution features with rich semantics. 
GAI can be integrated with any deep convolutional network for efficient semantic segmentation. 
In experiments, the GAI-based semantic segmentation networks, \ie, GAIN, can achieve $78.8$ mIoU with $22.3$ FPS on Cityscapes and $80.6$ mIoU with $64.5$ on CamVid using an NVIDIA 1080Ti GPU, which are the new state-of-the-art results of low-latency semantic segmentation.
Code and models are available at \url{https://github.com/hustvl/simpleseg}.}




\maketitle

\section{Introduction}
Recent research~\cite{WangSCJDZLMTWLX} has emphasized the importance of high-resolution semantic features to enhance the feature representations. Most methods \cite{WangSCJDZLMTWLX,NohHH15DeConvNet,abs-1708-04943StackDeconv,RonnebergerFB15Unet,hyperseg} construct high-resolution features through coordinate-based interpolation, \eg, bilinear interpolation, to upsample low-resolution semantic features to high-resolution for pixel-wise predictions.
Interpolation operators will resample features by geometric information such as distances and adjacency.
\begin{figure}[tbp]
    \centering
    \includegraphics[width=0.95\linewidth]{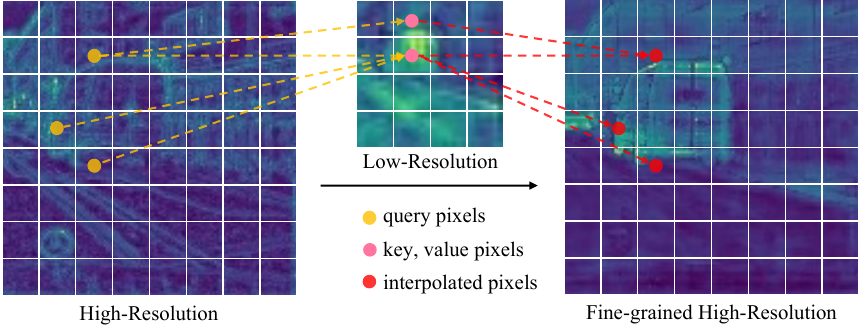}
    \caption{\textbf{Guided Attentive Interpolation.} GAI will build the pixel-level pairwise relations between \textit{query} points and \textit{key} points from high-resolution features and low-resolution features respectively, and leverage the relations to interpolate high-resolution semantic features.}
    \label{fig:comparison_interpolation}
\end{figure}
As for bilinear upsampling, each pixel will be interpolated by the weighted average of the surrounding pixels, and the weights will be decided by the geometric distances.
As for a parametric deconvolution~\cite{NohHH15DeConvNet}, features of a pixel will be aggregated with those of pixels in a fixed local region (\eg, $2\times2$) according to the learned weights.

However, these approaches for upsampling features are all based on coordinates and geometric constraints, while ignoring the semantic relations among pixels and fail to enrich the semantic information for high-resolution features.
In addition, due to several downsampling operations in the convolutional networks, traditional interpolation operators will cause feature misalignment between low-resolution and high-resolution features~\cite{LiYZZYYTT20SFNet,abs-2003-00872AlignSeg}, leading to wrong predictions.

To tackle these issues, we present a novel Guided Attentive Interpolation (GAI) module to interpolate features according to the pairwise spatial and semantic relations of all pixels based on the attention mechanism~\cite{VaswaniSPUJGKP17}, which can produce high-resolution, spatial-aligned, semantic-enriched,  context-enhanced features for pixel-level predictions.
Recently, attention mechanisms~\cite{0004GGH18,HuangWHHW019,ZhuXBHB19,FuLT0BFL19} have been widely explored to capture the long-range dependencies of pixels in semantic segmentation, in which image features are treated as \textit{query},~\textit{key}, and \textit{value} to calculate the pairwise relations and aggregate features.
We leverage the high-resolution features as the \textit{query}, which contains more spatial details to provide more guidance for aligning low-resolution features when upsampling.
And the low-resolution features, abundant in semantic information, act as the \textit{key} and \textit{value}, to provide semantic features for high-resolution feature maps by attention.
GAI can acquire the pairwise relations between pixels from high-resolution features and pixels from low-resolution features through a simple dot product. 
With the pairwise relations of pixels from different feature levels, attention can aggregate the features for each pixel by the weighted sum of other pixels.
As illustrated in Fig.~\ref{fig:comparison_interpolation}, GAI can interpolate the features according to the relations of all pixels instead of a local region as traditional interpolations.
It can both enrich the contextual information for high-resolution features and alleviate feature misalignment through pairwise relations with guidance from high-resolution features.
 In addition, GAI is adaptive to many attention modules, \eg, Non-local~\cite{0004GGH18} and Criss-Cross Attention~\cite{HuangWHHW019}. Considering the large computation budget and memory consumption of the standard spatial attention, we adopt Criss-Cross Attention as our basic attention module in this paper.

The state-of-the-art semantic segmentation methods tend to obtain contextual and high-resolution features by adopting backbones with dilated convolution~\cite{YuK15MultiScaleContext,ChenPKMY18,ChenPSA17,ChenZPSA18EncoderAtrous}, 
feature pyramid networks~\cite{LiYZZYYTT20SFNet}, 
or encoder-decoder networks~\cite{LinMSR17RefineNet,BadrinarayananK17SegNet,RonnebergerFB15Unet,NohHH15DeConvNet,ChenZPSA18EncoderAtrous}, 
which bring lots of computation cost and are infeasible for low-latency approaches. 
In this paper, we apply GAI to obtain high-resolution semantic features by aggregating high-resolution spatial features and low-resolution semantic features 
After extracting multi-scale features from the backbone network, \eg, ResNet~\cite{HeZRS16ResNet},
we employ the GAI to interpolate the semantic features of low resolution to a higher resolution, specifically $\frac{1}{8}\times$. In this way, GAI-based networks (GAIN) can acquire fine-grained semantic features with high resolution for accurate segmentation without heavy computation costs for building context modules and complex fusions.
Relying on the effective GAI modules, GAIN is rather compact and simple, thus achieving low-latency inference with high recognition accuracy. 

Finally, the main contribution of this paper can be summarized as follows:
\begin{itemize}
    \item We propose the Guided Attentive Interpolation method to produce high-resolution, spatial-aligned, semantic-enriched and context-enhanced deep feature maps. It is a novel and extremely effective feature upsampling operator that can be widely applied in deep learning. 
    \item We propose a compact and efficient semantic segmentation framework, GAIN, based on ResNet-18~\cite{HeZRS16ResNet} or DF-2~\cite{LiZPF19DFNet} and two Guided Attentive Interpolation modules.
    \item GAIN is fast and accurate: $78.8$ mIoU and $78.2$ mIoU on Cityscapes \cite{CordtsORREBFRS16} \textit{val} and \textit{test} respectively and can reach $22.3$ FPS with $1024\times2048$ input. In addition, GAIN achieves $80.6$ mIoU with $64.5$ FPS on CamVid~\cite{BrostowFC09Camvid} and outperforms most methods for real-time semantic segmentation. Moreover, we extend the real-time setting into ADE20K~\cite{ZhouADE20k} dataset and the GAIN achieves $39.1$ mIoU with $81.8$ FPS.
\end{itemize}

\section{Related Work}

\subsection{Semantic Segmentation}

Fully convolutional network (FCN)~\cite{LongSD15} has greatly promoted the development of semantic segmentation. Current research for high-quality segmentation can be divided into two groups, one of which focuses on gathering more contextual information for segmentation.
Zhao \etal exploits the pyramid pooling module \cite{ZhaoSQWJ17} to aggregate global context features from different levels for scene parsing. DeepLab and its improved versions~\cite{ChenPKMY18,ChenPSA17} adopt the dilated convolution to enlarge the respective fields and propose an Atrous Spatial Pyramid Pooling (ASPP) module to capture multi-scale context. \cite{YangYZLY18} presents a densely-connected ASPP module to generate multi-scale features to cover a large and dense range of scales. \cite{LiuRB15,0005DSZWTA18Encoding,YuWPGYS18DFN,HouZCF20Strip} exploit global average pooling to enlarge the receptive field and attain the global context. 

The other group tries to obtain the high-resolution fine-grained feature representations for more precise segmentation. Amounts of approaches~\cite{LinMSR17RefineNet,BadrinarayananK17SegNet,RonnebergerFB15Unet,NohHH15DeConvNet,ChenZPSA18EncoderAtrous,abs-1708-04943StackDeconv,ZhangZPXS18ExFuse} leverage a encoder-decoder style architecture to recover the high-resolution features with abundant semantic information. Wang~\etal proposes a high-resolution network (HRNet)~\cite{WangSCJDZLMTWLX} to maintain the high-resolution features and enhance the high-resolution representations by aggregating features from other resolutions.
PointRend~\cite{KirillovWHG20PointRend} adaptively sample key points with rich contextual information to obtain fine-grained features iteratively. 
CARAFE~\cite{WangCX0LL19CAFARE} addresses the lack of semantic information when upsampling features by interpolations and presents a content-aware deconvolution kernel to reassemble features. Though CARAFE can enlarge the receptive field and provide more context, the pixels for upsampling are still limited in a local region. 
\revise{Several works~\cite{hybridfeature,LiYZZYYTT20SFNet,abs-2003-00872AlignSeg} propose pixel-wise feature alignment modules by predicting the offsets and directions to alleviate the misalignment problems.
\cite{distillrs} enhances the high-resolution features by leveraging the super-resolution-assisted learning, which demonstrates a promising and effective direction.}
Differently, the proposed Guided Attentive Interpolation exploits the long-range semantic relations and can engage more semantic features when interpolating high-resolution features.


\subsection{Low-latency Semantic Segmentation}

Quantities of works have been chasing high-quality segmentation while research on low-latency semantic segmentation is also essential and enables many practical applications such as autonomous driving and scene understanding.
Zhao \etal exploits an image cascade network~\cite{ZhaoQSSJ18ICNet} to fuse features from multiple resolutions for efficient and accurate segmentation.
\cite{YuWPGYS18BiSeNetV1,abs-2004-02147BiSeNetV2} adopt a detail branch for high-resolution feature representations and a semantic branch for high-level contextual information.
DCNet~\cite{DCNet} adopts two independent networks for region-level and pixel-level context modeling and obtains good inference speeds.
ERFNet~\cite{RomeraABA18ERFNet} leverages residual connections and factorized convolutions to lower the latency and retain the segmentation accuracy.
ESPNet~\cite{MehtaRCSH18ESPNet} reduces the computation cost by substituting the combination of point-wise convolutions and spatial pyramid of dilated convolutions for normal convolution.
Houfi~\etal~\cite{HoufiM20} design efficient segmentation networks with shuffle/depthwise/grouped convolutions and achieve good inference speeds.
SFNet~\cite{LiYZZYYTT20SFNet} addresses the feature misalignment issue in feature pyramids and proposes a feature alignment module and an efficient network for fast semantic segmentation.
Recently, several methods~\cite{ChenGLZLW20FasterSeg,ZhangQLYLM19} adopting architecture search also perform well in terms of accuracy and latency.

\subsection{Attention Mechanism in Convolutional Neural Networks}
Attention mechanisms \cite{VaswaniSPUJGKP17}, especially self-attention, are widely adopted to obtain contextual information and enhance the feature representations in semantic segmentation \cite{0004GGH18,HuangWHHW019,ZhuXBHB19,0001XLWH19,0005ZWYLL19,YuLGSS20,FuLT0BFL19,YuWGYSS20ContextPrior,ZhaoZLSLLJ18PSA,0005DSZWTA18Encoding,YuanCW20OCRNet,FPANet}. 
Several works~\cite{HuangWHHW019,0001XLWH19,0005ZWYLL19,YuLGSS20} address the huge computation and memory consumption of Non-local blocks and propose efficient variations. 
Yu \etal propose an affinity loss~\cite{YuWGYSS20ContextPrior} to supervise the context learning for self-attention.
Zhu \etal present an asymmetric non-local block~\cite{ZhuXBHB19} to fuse multi-level features and regard the high-level features (stage-5) as \textit{query} and low-level features (stage-4) as \textit{key, value} respectively, and the asymmetric non-local block focuses more on the global context and neglect the local details due to the pyramid sampling. The proposed Guided Attentive Interpolation is completely different from \cite{ZhuXBHB19} and we propose to interpolate high-resolution semantic features through semantic and spatial relations and regard the high-resolution low-level features as \textit{query} and low-resolution high-level features as \textit{key, value}.

Recently, vision transformers~\cite{ViT} have made significant progress, and several methods~\cite{setr,segformer,segmenter,TopFormer,seaformer,semask,segvit,oneformer} adopt vision transformers for semantic segmentation.
SETR~\cite{setr} splits images into patches and feeds patches into a vision transformer to obtain the segmentation results through a convolutional decoder.
SegFormer~\cite{segformer} proposes a hierarchical vision transformer, \ie, MiT, for multi-scale fusion, and significantly improves semantic segmentation.
\revise{TopFormer~\cite{TopFormer} and SeaFormer~\cite{seaformer} design vision transformer architectures for mobile devices and obtain good results on mobile semantic segmentation.}
However, those works focus on the vision transformers in semantic segmentation and have achieved good performance. However, due to the quadratic computational cost of transformers, it is hard to achieve real-time inference speeds, especially for high-resolution images, \eg, $1024\times2048$.

\section{Our approach}
\subsection{Guided Attentive Interpolation}

In deep CNN, high-resolution features contain more spatial details but lack semantic information while low-resolution features encode more semantic and contextual information. In terms of semantic segmentation, we tend to fuse low-resolution and high-resolution features to keep both spatial and contextual information. 
Straightforwardly fusing features of different resolutions by traditional interpolation operators lacks semantic information and leads to feature misalignment, which is not friendly to pixel-wise prediction tasks, \eg, semantic segmentation. 

Motivated by that the self-attention mechanism can provide the pairwise relations among all pixels, we propose a Guided Attentive Interpolation to construct the mappings of pixels from high-resolution feature maps and low-resolution feature maps by the pairwise relations.
Through the relations between pixels in the low-resolution feature maps and that in the high-resolution feature maps, we can interpolate the low-resolution feature maps to high resolution by aggregating the features from all pixels, thus enriching more semantic information for high-resolution feature maps.

\begin{figure}[htbp]
    \centering
    \includegraphics[width=0.9\linewidth]{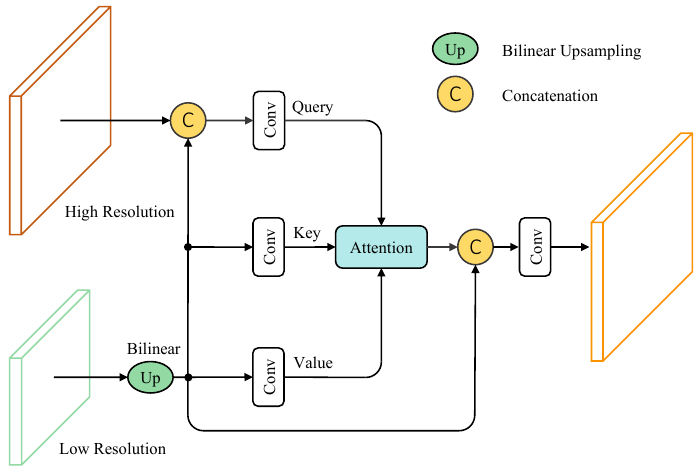}
    \caption{\textbf{Guided Attentive Interpolation Module.} The low-resolution feature maps will be interpolated to the same size as the high-resolution feature maps. The concatenation of high-resolution and low-resolution feature maps is defined as the \textit{query}. All $1\times1$ convolutions are used to reduce the dimension for less computation budget.}
    \label{fig:gai_module}
\end{figure}

As illustrated in Fig.~\ref{fig:gai_module}, Guided Attentive Interpolation aims at interpolating the low-resolution features to high resolution by leveraging the pairwise relations of pixels between high-resolution and low-resolution feature maps.
In Fig.~\ref{fig:gai_module}, the low-resolution features (green) will be interpolated to the same size as the high-resolution features (orange) which can be regarded as a coarse upsampling and provides semantic contexts for high-resolution features.
The interpolated low-resolution features (coarse high-resolution features) termed as $F_l$ and the high-resolution features termed as $F_h$ will be concatenated as the \textit{query} features.
The concatenation is crucial in Guided Attentive Interpolation since it can simultaneously provide spatial information to align low-resolution features and also bring much contextual information by self-attention. 

In Guided Attentive Interpolation, the \textit{query} features input to attention is defined as follows:
\begin{equation}
Q = \text{conv}([F_h, F_l]) \in \mathbb{R}^{C\times H \times W},
\end{equation}
where we use a $1\times1$ convolution to reduce the feature dimension after concatenation.

Then $F_l$ will also be used as the \textit{key} $K \in \mathbb{R}^{C\times H' \times W'} $  and \textit{value} $V \in \mathbb{R}^{C\times H' \times W'}$ to calculate the relations of pixels from the same level or the higher level by follows:
\begin{equation}
    A = \text{Softmax}(f(W^QQ, W^KK)),
\end{equation}
where $f$ is the affinity function to calculate the affinity matrix $A$. $W^Q \in \mathbb{R}^{d_k\times C}$ and $W^K \in \mathbb{R}^{d_k\times C}$ are $1\times1$ projection convolutions without non-linearity. $C$ and $d_k$ is set to $128$ and $8$ respectively for lower computation cost.
In the original attention~\cite{VaswaniSPUJGKP17}, $f$ is a simple dot-product operation and \revise{$A \in \mathbb{R}^{HW\times H'W'}$, where the query has the size of $H\times W$ and the key has the size of $H'\times W'$.}
When using Criss-Cross Attention, $f$ will calculate the dot-product along the horizontal and vertical direction and $A \in \mathbb{R}^{H\times W\times (H'+W'-1)}$ for less computation budget. After obtaining the affinity between \textit{query} and \textit{key}, the new output of the attention can be formulated as follows:
\begin{equation}
    O_p = \sum^N_i A_{p,i} \cdot (W^VV_{p,i}),  
\end{equation}
where $p$ denotes the $p$-th pixel in the feature map, $A_{p,i}$ and $V_{p,i}$ denote the affinity weight and feature vector of the $i$-th pixel which is used to update the features of $p$-th pixel. $W^V \in \mathbb{R}^{d_v\times C}$ is a $1\times1$ projection convolution in which $d_v$ is set to $64$. Using Criss-Cross Attention, $N$ is set to $H'+W'-1$ because a pixel will be updated by the pixels along the horizontal and vertical directions.
\revise{Therefore, the computation complexity is $\mathbf{O}(HW(H'+W'-1))$, which is significantly reduced compared to the standard attention ($\mathbf{O}(HWH'W')$).}
The output features will be concatenated with the original features and then output by a $1\times1$ convolution.

Compared with traditional interpolation operators, the Guided Attentive Interpolation can be regarded as a fine-grained upsampling which brings more semantic information for pixels in high-resolution feature maps.
In addition, the basic attention module is general and can be replaced with various other attention modules.
Using Criss-Cross Attention can largely reduce computation and memory budget. 
Furthermore, reducing the dimension of features for attention can further lower the cost with little performance degradation.

\subsection{Guided Attentive Interpolation for Semantic Segmentation}
\begin{figure*}
    \centering
    \includegraphics[width=\linewidth]{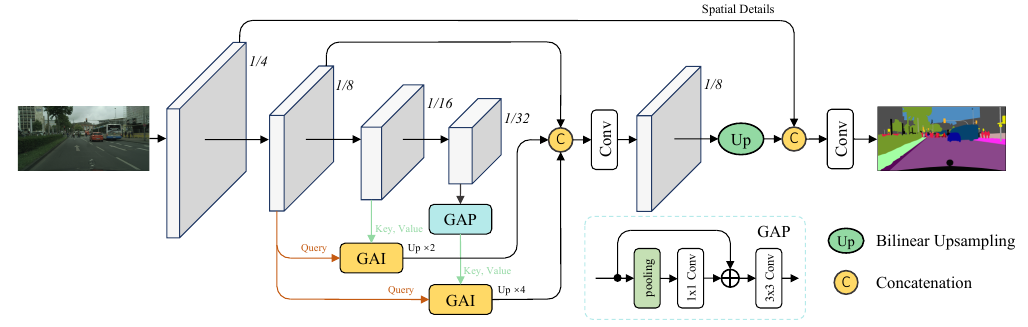}
    \caption{The network architecture of our proposed \textbf{GAIN (GAI-based Network)} with two Guided Attentive Interpolation modules to interpolate features from \{C4, C5\} of ResNet (or DF-2) to $\times\frac{1}{8}$ resolution for fusion as the fine-grained semantic features. GAP denotes the global average pooling. All convolutions are $1\times1$ for less computation budget.}
    \label{fig:main_arch}
\end{figure*}

With the advantages of Guided Attentive Interpolation for efficiently aggregating high-resolution with rich semantics, we design an efficient segmentation network, namely GAIN (GAI-based Network), to deal with the multi-scale features from the backbone.
As shown in Fig.~\ref{fig:main_arch}, we use ResNet-18~\cite{HeZRS16ResNet} (or DF-2~\cite{LiZPF19DFNet}) as the backbone for feature representations.
The proposed GAIN utilizes multi-scale feature maps \{$\frac{1}{4}$, $\frac{1}{8}$, $\frac{1}{16}$, $\frac{1}{32}$\}(\{C2, C3, C4, C5\}) from the backbone for contexts of different scales.
\revise{Considering that using $\frac{1}{4}\times$-resolution features will increase computational overhead, we mainly adopt $\frac{1}{8}\times$-resolution features as our default high-resolution features.}
As for the two low-resolution feature maps (\{$\frac{1}{16}$, $\frac{1}{32}$\}), we apply two GAI modules to interpolate the features to a higher resolution $\frac{1}{8}\times$ and then concatenate all the $\frac{1}{8}\times$-resolution feature maps with a following $1\times1$ convolution for dimension reduction and a $3\times3$ convolution for spatial feature fusion.
\revise{
Considering that the image features of C5 ($\frac{1}{32}\times$) which contains more semantic contexts than low-level features such as C2 or C3, we insert a simple but effective global average pooling module (GAP shown in Fig.~\ref{fig:main_arch}) after the lowest-resolution features (C5) before the GAI module to further enhance global context of image features. 
Then the high-level enhanced image features will be interpolated to higher-resolution features through the proposed GAI. Therefore, we can obtain high-resolution and semantic-enriched features for image segmentation.}
Considering that using $\frac{1}{4}$ features brings a huge computation burden while leading to minor improvements compared to using $\frac{1}{8}$ features, we only apply the proposed Guided Attentive Interpolation modules for $\frac{1}{8}$ features.
The Guided Attentive Interpolation modules bring rich semantic information for the $\frac{1}{8}$ feature maps.
For more precise segmentation, we directly employ the $\frac{1}{4}$-resolution feature maps (C2) from ResNet to provide more spatial details.
At last, a $1\times1$ convolution classifier will take the spatial features and the high-resolution context features and then output the final segmentation results.

\section{Experiments}

We perform extensive experiments on the Cityscapes dataset, CamVid dataset, and ADE20K dataset to evaluate both the segmentation accuracy and inference speed of our proposed GAIN and demonstrate the effectiveness of the Guided Attentive Interpolation with ablation experiments.

\subsection{Datasets and Evaluation Metrics}

\textbf{Cityscapes.} Cityscapes is a large urban scene parsing dataset, 
containing $19$ categories and 5,000 high-quality annotated($1024\times2048$) images for street view scene segmentation, 
in which 2,975 images are used for training, 500 and 1525 images for validation and testing respectively. 
In our experiments, we only use fine-annotated images for training and testing. 

\textbf{CamVid.} Cambridge-driving Labeled Video Database (CamVid)~\cite{BrostowFC09Camvid} is a driving scene dataset which contains 701 images with $720\times960$ resolution extracted from the video sequences. 
The images are split into 367 training, 101 validation, and 233 testing images and labeled for $11$ categories for segmentation.

\textbf{ADE20K.} ADE20K~\cite{ZhouADE20k} is a challenging scene parsing dataset which contains 20,210 training images with $150$ semantic categories and 2,000 and 3,352 images for validation and testing respectively.

\revise{\textbf{PASCAL Context.} PASCAL Context~\cite{pascalcontext} extends the PASCAL VOC~\cite{pascalvoc} dataset by annotating amounts of object and background classes for segmentation, which contains 4998 images for training and 5015 images for validation. The PASCAL Context dataset contains 59 categories.}


\revise{All models are trained on the training set and evaluated on the validation or test set.}
\subsection{Implementation Details}

Our model is developed based on the PyTorch framework. 
We adopt ResNet-18~\cite{HeZRS16ResNet} and DF-2~\cite{LiZPF19DFNet} pre-trained on ImageNet as our backbone networks and other parameters are randomly initialized.

\textbf{Data Augmentation.} 
As for training, we apply the random horizontal flipping and random scaling from $0.5$ to $2.0$ and then randomly crop the image to a fixed size $1024\times1024$ for Cityscapes, $720\times960$ for CamVid, and $512\times512$ for ADE20K \revise{and PASCAL Context}. Color jittering including brightness, contrast, saturation, and hue is adopted. 
In the inference phase, we adopt the original size $1024\times2048$ and $720\times960$ for Cityscapes and CamVid without any augmentations. Considering the variable sizes of images from ADE20K \revise{and PASCAL Context}, we resize each image to have a shorter side of $512$ and pad the longer side to be multiple of $32$.

\revise{
\textbf{Metrics.}
We mainly adopt \textbf{mIoU} (mean intersection over union ) to evaluate segmentation accuracy, which measures the overlap between the predicted segmentation and the ground-truth segmentation.
}

\textbf{Training.} 
Following the common practice~\cite{LiYZZYYTT20SFNet,abs-2003-00872AlignSeg}, we train all models using Synchronized SGD to optimize with Synchronized Batch Normalization and 16 images per batch on 8 NVIDIA 2080Ti GPUs. During training, the learning rate is decayed according to the `poly' learning rate strategy: $ lr = lr_0*(1-\frac{iter}{max\_iter})^p$ ($p = 0.9$), the initial learning rate $lr_0$ is set as $0.01$. \revise{Models are trained $80$K, $5$K, $160$K, and $100$K iterations for Cityscapes, CamVid, ADE20K, and PASCAL Context respectively.}

\textbf{Auxiliary Supervision.}
To strengthen the feature representation of intermediate features and boost \revise{the network} optimization, we append auxiliary heads after the outputs of two Guided Attentive Interpolation modules to generate intermediate segmentation results. The auxiliary heads are simple and consist of two convolutions with Batch Normalization.
Further, we adopt auxiliary loss to supervise the intermediate outputs. 
Ultimately, the total loss of our proposed method is defined as follows:
\begin{equation}
    \mathcal{L} = \mathcal{L}_{out} + \lambda\cdot\mathcal{L}_{aux},
\end{equation}
where $\mathcal{L}_{out}$ and $\mathcal{L}_{aux}$ are adopted to supervise the final outputs and intermediate outputs of the GAI. $\lambda$ is set to $1.0$ in all experiments.
\revise{We adopt cross-entropy loss equipped with online hard example mining for both two losses, which is defined as follows: 
\begin{equation}
    \mathcal{L}(p_{x,y}, q_{x,y}) = -\sum^C_iy^i_{x,y}\log(p^i_{x,y}),
\end{equation}
where $p_{x,y}$ and $q_{x,y}$ are the prediction and ground-truth label for the pixel $(x, y)$ and $C$ is the number of the classes.}

\textbf{Inference.}
As for inference, we input the original image to our model without any augmentation such as multi-scale inputs or horizontal flipping. Unless specified, we input a single image on a single NVIDIA 1080Ti GPU  to measure the inference speed. TensorRT and mixed-precision is not adopted for further acceleration in our implementation.

\subsection{Results on Cityscapes}
\begin{figure}
    \centering
    \includegraphics[width=0.90\linewidth]{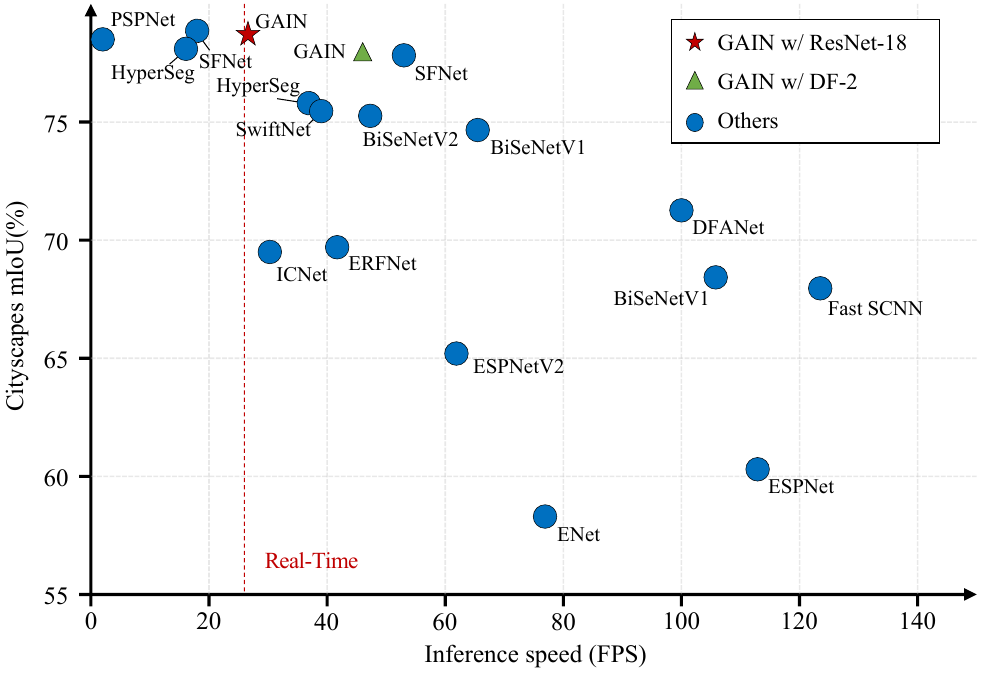}
    \caption{\textbf{Speed-accuracy trade-off.} Our methods are presented in stars and triangles for different backbones. Other methods are presented in blue circles. Our proposed GAIN achieves a superior trade-off between speed and segmentation accuracy.}
    \label{fig:speed_accuracy}
\end{figure}

In this section, we compare the proposed GAIN with other state-of-the-art methods on the Cityscapes dataset. As shown in Tab.~\ref{tab:main_results}, GAIN can achieve strong segmentation results with fast speed, \ie, $22.3$ FPS for ResNet-18 and $43.8$ FPS for DF-2 with $1024\times2048$ input.
\revise{Compared to transformer-based methods, such as PEM~\cite{pem} and SegNeXt~\cite{segnext}, the proposed GAIN can achieve superior inference speed even with high-resolution inputs.}
Besides, GAIN has much simple and compact structures and benefits much from Guided Attentive Interpolation modules for fine-grained high-resolution features.
The Guided Attentive Interpolation can capture the local contextual information for fine-grained details and large contextual information for semantics with no need for extra designs for spatial features and semantic features.
Fig.~\ref{fig:speed_accuracy} illustrates the speed-and-accuracy trade-off on the Cityscapes dataset.

\begin{table*}[htbp]
    \centering
    \caption{\textbf{Comparison with state-of-the-art methods on Cityscapes.} We evaluate our proposed GAIN with $1024\times2048$ input on Cityscapes \textit{val} and \textit{test}. $^\dag$ indicates that the method is accelerated by TensorRT. Inference speeds are measured on one NVIDIA 1080 Ti.} \par
    \small
    \renewcommand\arraystretch{1.2}
    {
    \begin{tabular}{l|cc|cc|c}
        \toprule
        \multirow{2}{*}{Method} & \multirow{2}{*}{Backbone} & \multirow{2}{*}{Resolution} & \multicolumn{2}{c|}{mIoU(\%)} & \multirow{2}{*}{FPS} \\ 
         & & & \textit{val} & \textit{test} \\
        \hline

        \hline
        ENet~\cite{PaszkeCKC16ENet} & -  & $640\times360$ & - & $58.3$  & $76.9$\\
        ESPNet~\cite{MehtaRCSH18ESPNet} & - & $512\times1024$ & - & $60.3$  & $112.9$\\
        ESPNetV2~\cite{MehtaRCSH18ESPNet} & - & $512\times1024$ & $66.4$ & $65.2$ & $61.9$\\
        ERFNet~\cite{RomeraABA18ERFNet} & - & $512\times1024$ & - & $69.7$ & $41.7$\\
        ICNet~\cite{ZhaoQSSJ18ICNet} & ResNet-50 & $1024\times2048$ & - & $69.5$ & $30.3$ \\
        Fast SCNN~\cite{PoudelLC19} & - & $1024\times2048$ & $68.6$ & $68.0$ & $123.5$ \\
        DFANet~\cite{LiXFS19} & Xception-A & $1024\times1024$ & - & $71.3$ & $100.0$\\
        \revise{MLFNet~\cite{mlfnet}} & \revise{ResNet-34} & \revise{$512\times1024$} & \revise{-} & \revise{72.1} & \revise{72.0} \\
        \revise{LETNet~\cite{LETNe}} & \revise{-} & \revise{$512\times1024$} & \revise{-} & \revise{72.8} & \revise{90.5} \\
        SwiftNet~\cite{OrsicKBS19} & ResNet-18 & $1024\times2048$ & $75.4$ & $75.5$ & $39.9$\\
        SFNet~\cite{LiYZZYYTT20SFNet} & DF-2 & $1024\times2048$ & - & $77.8$ & $53.0$ \\
        SFNet~\cite{LiYZZYYTT20SFNet} & ResNet-18 & $1024\times2048$ & $78.3$ & $78.9$ & $18.0$ \\
        BiSeNetV1~\cite{YuWPGYS18BiSeNetV1} & Xception-39& $768\times1536$ & $69.0$ & $68.4$ & $105.8$\\
        BiSeNetV1~\cite{YuWPGYS18BiSeNetV1} & ResNet-18 & $768\times1536$ & $74.8$ & $74.7$ & $65.5$\\
        BiSeNetV2$^\dag$~\cite{abs-2004-02147BiSeNetV2} & - & $512\times1024$ & $75.8$ & $75.3$ & $47.3$ \\
        HyperSeg-M~\cite{hyperseg} & EfficientNet-B1 & $512\times1024$ & $76.2$ & $75.8$ & $36.9$ \\
        HyperSeg-S~\cite{hyperseg} & EfficientNet-B1 & $768\times1536$ & $78.2$ & $78.1$ & $16.1$ \\
        STDC1$^\dag$~\cite{STDC} & STDC1 & $768\times1536$ & $74.5$ & $75.3$ & $126.7$ \\
        STDC2$^\dag$~\cite{STDC} & STDC2 & $768\times1536$ & $77.0$ & $76.8$ & $97.0$ \\
        \revise{SegNeXt~\cite{segnext}} & \revise{-} & \revise{$768\times1536$} & \revise{-} &  \revise{$78.0$} & \revise{12.6} \\
        \revise{PEM~\cite{pem}} & \revise{STDC1} & \revise{$1024\times2048$} & \revise{78.3} & \revise{-} & \revise{16.6} \\
        \revise{PEM~\cite{pem}} & \revise{STDC2} & \revise{$1024\times2048$} & \revise{79.0} & \revise{-} & \revise{14.2} \\
        \revise{RDRNet~\cite{RDRNet}} & \revise{-} & \revise{$1024\times2048$} & \revise{78.9} & \revise{78.3} & \revise{24.2} \\
        \revise{BiDGANet-B~\cite{Liao2023BilateralNW}} & \revise{-} & \revise{$1024\times2048$} &  \revise{75.2} & \revise{-} &\revise{39.8} \\
        \revise{BiDGANet-L~\cite{Liao2023BilateralNW}} & \revise{-} & \revise{$1024\times2048$} & \revise{77.9} & \revise{-} & \revise{23.5} \\
        \hline
        GAIN & DF-2 & $1024\times2048$ & $78.3$ & $77.9$ & $43.8$\\
        GAIN & ResNet-18 & $1024\times2048$  & $78.8$ & $78.2$ & $22.3$ \\
        \bottomrule
    \end{tabular}}
    \label{tab:main_results}
\end{table*}

\subsection{Results on CamVid}
Tab.~\ref{tab:camvid_main_results} shows the comparisons with the state-of-art methods on CamVid dataset. Our proposed GAIN with DF-2 achieves $74.2$ mIoU  and $92.3$ FPS with the input size $720\times960$, which is superior than previous methods in terms of the trade-off between speed and accuracy. With Cityscapes pre-trained weights, GAIN achieves $80.6$ mIoU and $64.5$ FPS on CamVid.

\begin{table*}[htbp]
    \centering
    \caption{\textbf{Comparison with state-of-the-art methods on CamVid.} We evaluate our proposed GAIN with $720\times960$ input on CamVid \textit{test}. $^\dag$ indicates that the method is accelerated by TensorRT. Inference speeds are measured on one NVIDIA 1080 Ti. $^\S$ denotes GAIN using Cityscapes pre-trained weights.} \par
    \small
    \renewcommand\arraystretch{1.2}
    \begin{tabular}{l|cc|c|c}
        \toprule
        Method & Backbone & Resolution & mIoU(\%) & FPS \\
        \hline
        
        \hline
        ENet~\cite{PaszkeCKC16ENet} & - & $720\times960$ & $51.3$ & $61.2$ \\
        DFANet~\cite{LiXFS19} & Xception-B &  $720\times960$ & $64.7$ & $120$ \\
        ICNet~\cite{ZhaoQSSJ18ICNet} & ResNet-50 & $720\times960$ & $67.1$ & $34.5$ \\
        SwiftNet & ResNet-18 & $720\times960$ & $72.6$ & $-$ \\
        BiSeNetV1~\cite{YuWPGYS18BiSeNetV1} & Xception-39 & $720\times960$ & $65.6$ & $175$ \\
        BiSeNetV1~\cite{YuWPGYS18BiSeNetV1} & ResNet-18 & $720\times960$ & $68.7$ & $116.3$ \\
        \revise{MLFNet~\cite{mlfnet}} & \revise{ResNet-34} & \revise{$720\times960$} & \revise{69.0} & \revise{57.0} \\
        \revise{LETNet~\cite{LETNe}} & \revise{-} & \revise{$360\times480$} & \revise{70.5} & \revise{126.7} \\
        BiSeNetV2$^\dag$~\cite{abs-2004-02147BiSeNetV2} & - & $720\times960$ & $72.4$ & $124.5$ \\
        BiSeNetV2-L$^\dag$~\cite{abs-2004-02147BiSeNetV2} & - & $720\times960$ & $73.2$ & $32.7$ \\
        SFNet~\cite{LiYZZYYTT20SFNet} & DF-2 & $720\times960$ & $70.4$  & $134.1$ \\
        SFNet~\cite{LiYZZYYTT20SFNet} & ResNet-18 & $720\times960$ & $73.8$  & $35.5$ \\
        STDC1$^\dag$~\cite{STDC} & STDC1 & $720\times960$ & $73.0$ & $197.6$ \\
        STDC2$^\dag$~\cite{STDC} & STDC2 & $720\times960$ & $73.9$ & $152.2$ \\
        HyperSeg-M~\cite{hyperseg} & EfficientNet-B1 & $720\times960$ & $78.4$ & $38.0$ \\
        HyperSeg-L~\cite{hyperseg} & EfficientNet-B1 & $720\times960$ & $79.1$ & $16.6$ \\
        \revise{RDRNet~\cite{RDRNet}} & \revise{-} & \revise{$720\times960$} &  \revise{78.4} & \revise{49.2} \\
        \hline
        GAIN & DF-2 & $720\times960$ & $74.2$ &  $92.3$\\
        GAIN & ResNet-18 & $720\times960$ & $74.6$ & $64.5$ \\
        GAIN$^\S$ & ResNet-18 & $720\times960$ & $80.6$ & $64.5$ \\
        \bottomrule
    \end{tabular}
    \label{tab:camvid_main_results}
\end{table*}

\subsection{Results on ADE20K}
Since it is the first work to deal with real-time segmentation for ADE20K dataset, we re-implement PSPNet~\cite{ZhaoSQWJ17}, BiSeNetV1~\cite{YuWPGYS18BiSeNetV1}, and SFNet~\cite{LiYZZYYTT20SFNet} according to their official code for comparison. 
All models are trained with the same setting. 
\revise{Tab.~\ref{tab:ade_main_results} shows the comparisons with state-of-the-art methods on the ADE20K dataset. GAIN with ResNet-18 can achieves comparable accuracy but faster inference speed compared to SFNet and PSPNet. In addition, we further compare the computation cost (FLOPs) and parameters in Tab.~\ref{tab:ade_main_results}.
Specifically, we adopt the $512\times512$ as the input resolution and calculate the FLOPs.
Tab.~\ref{tab:ade_main_results} shows that the proposed GAIN has fewer parameters than the previous methods, such as SFNet. In addition, the backbone, \ie, ResNet-18, contains 11.3M parameters, which is nearly 94\% of the whole model.
}

\begin{table*}
    \color{black}
    \centering
    \caption{\textbf{Comparison with state-of-the-art methods on ADE20K.} We evaluate the proposed GAIN on ADE20K \textit{val}. In addition, we compare the parameters and FLOPs among different methods.} \par
    \renewcommand\arraystretch{1.2}
    \small
    \begin{tabular}{l|ll|cc|cc}
    \toprule
    Method & Backbone & Resolution & Parameters & FLOPs & mIoU & FPS \\
    \midrule
    PSPNet~\cite{ZhaoSQWJ17} & ResNet-18 & $512\times$ & 14.5M & 67.72G & 38.90 & 52.5 \\
    BiSeNetV1~\cite{YuWPGYS18BiSeNetV1} & ResNet-18 & $512\times$ & 13.0M & 20.59G & 35.78 & 117.2 \\
    SFNet~\cite{LiYZZYYTT20SFNet} & ResNet-18 &  $512\times$ & 12.9M & 30.63G & 38.68 & 69.3 \\
    \hline 
    GAIN & ResNet-18 &  $512\times$ & 12.0M & 29.44G & 39.12 & 81.8 \\
    \bottomrule
    \end{tabular}
    \label{tab:ade_main_results}
\end{table*}

\revise{
\subsection{Results on PASCAL Context}
Tab.~\ref{tab:pascal_context} shows the experimental results on the PASCAL Context dataset, which can demonstrate the superior performance of the proposed GAIN in terms of both the segmentation accuracy and the inference speed.
}
\begin{table}
    \color{black}
    \centering
    \caption{Results on PASCAL Context \textit{val}. The results of the other methods are produced by their open-source code. We evaluate the FPS on the same machine with one NVIDIA 1080Ti GPU.}
    \setlength{\tabcolsep}{10pt}
    \begin{tabular}{l|l|cc}
    \toprule
     Method & Backbone & mIoU & FPS \\
     \midrule
     PSPNet~\cite{ZhaoSQWJ17} & ResNet-18 & 45.91 & 40.3 \\
     PSPNet~\cite{ZhaoSQWJ17} & DF-2 & 46.29 & 54.5 \\
     BiSeNet~\cite{abs-2004-02147BiSeNetV2} & ResNet-18 & 42.60 & 98.9 \\
     BiSeNet~\cite{abs-2004-02147BiSeNetV2} & DF-2 & 44.35 & 60.1 \\
     SFNet~\cite{LiYZZYYTT20SFNet} & ResNet-18 & 42.34 & 51.1 \\
     SFNet~\cite{LiYZZYYTT20SFNet} & DF-2 & 45.52 & 54.2 \\
     \hline
     Ours & ResNet-18 & 44.81 & 58.5 \\
     Ours & DF-2 & 47.48 & 61.2 \\
     \bottomrule
    \end{tabular}
    \label{tab:pascal_context}
\end{table}

\subsection{Ablation Experiments}
\textbf{Component Analysis in GAIN.}
We conduct ablative experiments to understand the key components of the proposed GAIN. 
The proposed GAIN is built on the backbone network, \ie, ResNet-18, with vanilla fusions from low-resolution features C4 and C5 to high-resolution features C3 through bilinear interpolation.
The stride of the fused features is $8$ and the final segmentation results are upsampled to $1024\times2048$. 
In Tab.~\ref{tab:analysis_components}, our baseline can reach $75.2$ mIoU on Cityscapes \textit{val} with the inference speed of $28.9$ ms per image. 
Then we (1) apply Guided Attentive Interpolation modules to replace the bilinear interpolation to interpolate high-resolution features from low-resolution features C4 and C5;
(2) exploit the auxiliary loss to supervise the intermediate outputs of the GAI modules.
(3) fuse $\frac{1}{4}\times$-resolution feature maps to attain more spatial details;
(4) adopt a simple global average pooling module to enlarge the receptive field.
Tab.~\ref{tab:analysis_components} indicates that our proposed Guided Attentive Interpolation can significantly improve the performance of Cityscapes semantic segmentation by $1.8$ mIoU.
Adding the auxiliary loss can straightforwardly contribute to the capability of the Guided Attentive Interpolation modules to obtain accurate pairwise relations, thus promoting better feature aggregations.
Utilizing the higher-resolution features with spatial details will further boost the performance and the extra global average pooling can obtain a $0.5$ mIoU gain with a negligible time cost. 
\begin{table*}
    \centering
    \caption{\textbf{Components in GAIN.} We evaluate the effectiveness of each component in GAIN step by step. Notations: GAI means adding Guided Attentive Interpolation, Auxilary, Spatial, and GAP denote the auxiliary supervision, spatial details, and Global Average Pooling, respectively.}
    \centering
    \small
    \renewcommand\arraystretch{1.2}
    \begin{tabular}{cccc|cc}
    \toprule
    GAI & Auxilary & Spatial & GAP  & mIoU(\%) & Time(ms) \\
    \hline
    
    \hline
      &  &  &  & $75.2$ & $28.9$ \\
    \cmark &  &  & & $77.0$ & $34.6$ \\
    \cmark & \cmark &  & & $77.6$ & $34.6$ \\
    \cmark & \cmark  & \cmark  & & $78.3$ & $37.5$\\
    \cmark & \cmark  & \cmark  & \cmark  & $78.8$ & $37.7$ \\
    \bottomrule
    \end{tabular}
    \label{tab:analysis_components}
\end{table*}

\begin{figure*}
    \centering
    \includegraphics[width=0.95\linewidth]{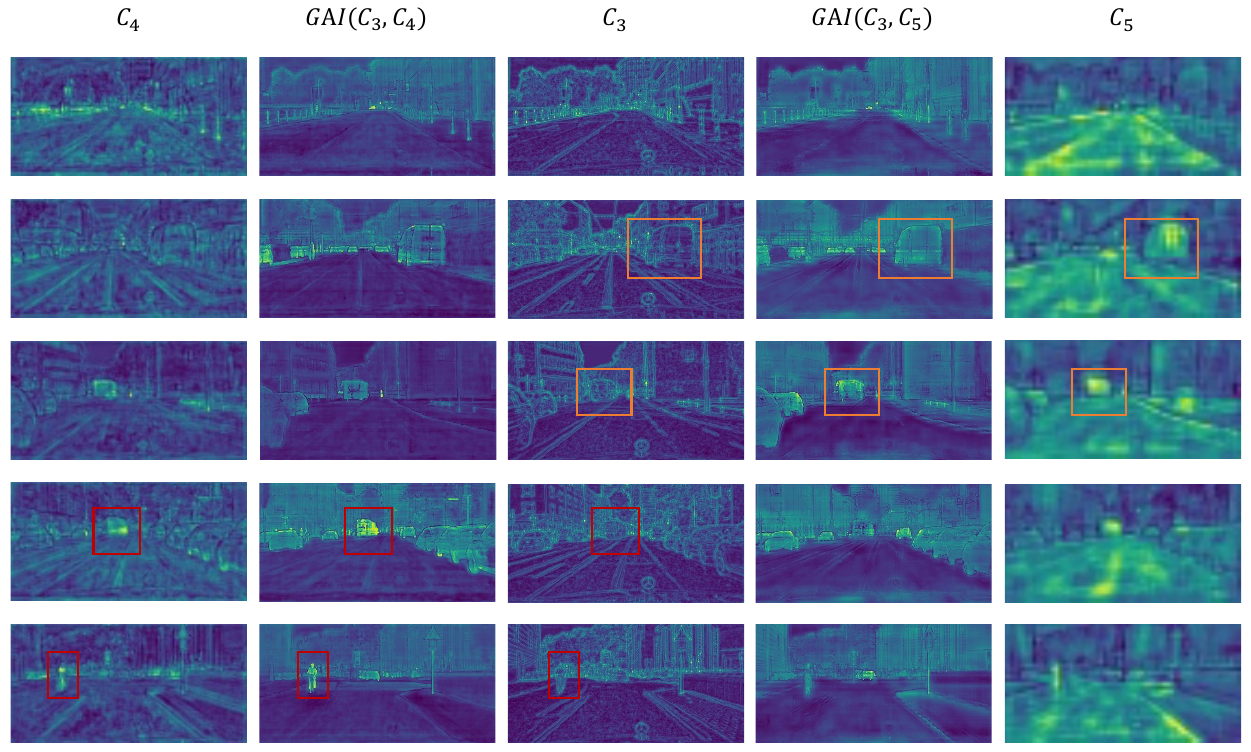}
    \caption{\textbf{Visualizations of feature maps before/after Guided Attentive Interpolation modules.} \{C3, C4, C5\} are the output features from different stages of the backbone (C3 contains higher-resolution feature maps). For feature visualization, we perform an element-wise sum along the channel axis for each $C$-channel features to obtain a single-channel feature map. The lighter area has a higher response. }
    \label{fig:attention_features}
\end{figure*}

\begin{figure}
    \centering
    \includegraphics[width=0.95\linewidth]{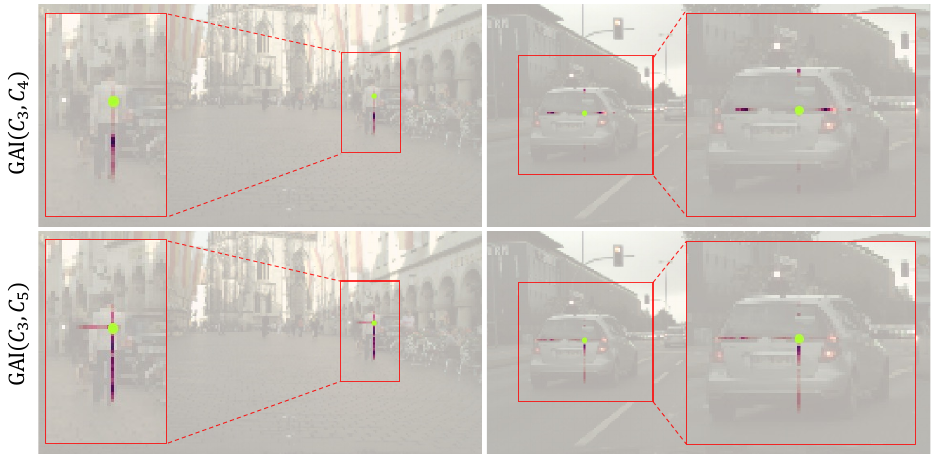}
    \caption{\textbf{Visualizations of Attention Weights.} The green points are the \textit{query} pixels of high-resolution features and the highlighted pixels in the maps are the \textit{key},
    \textit{value} pixels in low-resolution features. 
    The query pixels adaptively highlight the surrounding pixels to interpolate fine-grained high-resolution features by considering the semantic relations.
    The attention maps illustrate cross-shaped attention weights due to the use of Criss-Cross Attention.}
    \label{fig:attention_weights}
\end{figure}

The Guided Attentive Interpolation can explore the pixel-wise relations between low-resolution features and high-resolution features and interpolate new high-resolution feature maps by aggregating semantic features according to the relations.
Fig.~\ref{fig:attention_features} illustrates the visualizations of the features before/after using the Guided Attentive Interpolation.
The low-resolution C4 and C5 lack spatial information but with much contextual information while the high-resolution C3 contains more spatial information such as the redundant details that are necessary to exist in the final segmentation results.
From Fig.~\ref{fig:attention_features}, we observe that low-resolution feature maps are extremely coarse while high-resolution maps are fine. 
After the Guided Attentive Interpolation, fine-grained feature maps with rich semantics are generated, thus leading to better segmentation results. 
Fig.~\ref{fig:attention_weights} shows the attention weights of the given \textit{query} pixels (green color) from high-resolution features. The \textit{query} pixels tend to highlight the surrounding pixels (from low-resolution) which are semantically and spatially similar pixels.
Therefore, the proposed Guided Attentive Interpolation can interpolate high-resolution features with the consideration of both semantic and spatial relations, thus boosting the performance for dense prediction tasks.

\textbf{Comparisons of Feature Upsampling Methods.}
To validate the effectiveness of our proposed Guided Attentive Interpolation, we perform experiments with several different approaches, \ie, bilinear interpolation, CARAFE~\cite{WangCX0LL19CAFARE}, and FAM~\cite{LiYZZYYTT20SFNet} for aggregating high-resolution features. We only replace the Guided Attentive Interpolation modules with other approaches and keep other settings consistent with our proposed GAIN.
As shown in Tab.~\ref{tab:comparison_up}, the improvements brought by CARAFE and FAM are negligible compared to bilinear interpolation. Our proposed Guided Attentive Interpolation outperforms these methods by significant large margins, which can be attributed to that Guided Attentive Interpolation leverages the semantic relations of pixels to align features of different resolutions and gather more contextual information.

\begin{figure*}
    \centering
    \includegraphics[width=1.0\linewidth]{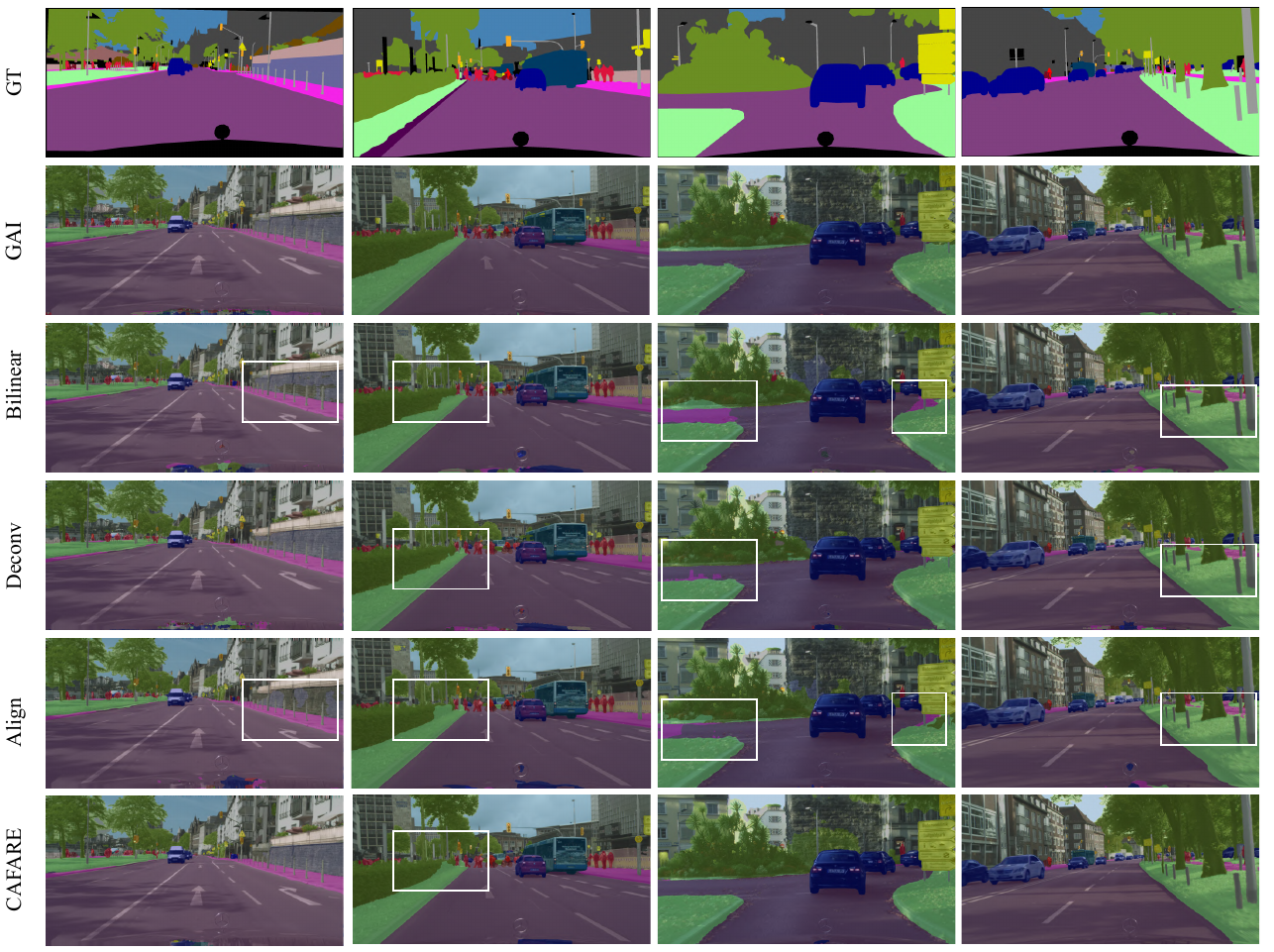}
    \caption{Qualitative results on the Cityscapes validation dataset of adopting different feature upsampling methods. GT denotes the ground-truth segmentation. \textbf{White Boxes} highlight some false predictions for comparison. It's clear that models with Guided Attentive Interpolation tend to produce higher-quality segmentation results.}
    \label{fig:qualititative}
\end{figure*}

Fig.~\ref{fig:qualititative} presents the qualitative results of different upsampling methods. Our proposed Guided Attentive Interpolation can achieve higher-quality segmentation results compared to other approaches.
The contextual information brought by the GAI module can reduce the probability of inter-class misclassification.
 
\begin{table}
    \caption{\textbf{Feature Upsampling Methods.} We replace the Guided Attentive Interpolation with other interpolation operations, \ie, bilinear interpolation, FAM~\cite{LiYZZYYTT20SFNet}, and CARAFE~\cite{WangCX0LL19CAFARE}. FLOPs for bilinear interpolation and grid sample (used in \cite{LiYZZYYTT20SFNet}) are ignored.}
    \centering
    \small
    \renewcommand\arraystretch{1.2}
    \begin{tabular}{l|ccc}
        \toprule
        Method & FLOPs($\Delta$) & mIoU(\%) & Time(ms)\\
        \hline
        
        \hline
        Bilinear & $0.0$G & $77.3$ & $32.4$\\
        CARAFE & $2.48$G & $77.6$ & $33.5$ \\
        FAM & $4.93$G & $77.3$ & $37.0$ \\
        \hline
        GAI & $9.95$G & $78.8$ & $37.7$ \\
        \bottomrule
    \end{tabular}
    \label{tab:comparison_up}
\end{table}

\textbf{Comparisons of Attention Module.}
To verify the effectiveness of our Guided Attentive Interpolation with other attention modules, we replace the Recurrent Criss-Cross Attention (RCCA) with Non-local. Tab.~\ref{tab:comparison_attention} shows the performance and speed of using Non-local and proves the effects of the Guided Attentive Interpolation. 
Due to the heavy computation of Non-local, the latency of the model rapidly increases.
Considering the speed and accuracy, we adopt Criss-Cross Attention as our basic attention module.

\begin{table}
\caption{\textbf{Attention Module.} Comparison of using different basic attention.}
\centering
\small
\renewcommand\arraystretch{1.2}
\begin{tabular}{l|ccc}
    \toprule
    Attention & FLOPs(G) & mIoU(\%) & Time(ms) \\
    \hline
    
    \hline
    Non-local & $310.59$ & $77.9$ & $194.4$ \\
    \hline
    RCCA & $9.95$& $78.8$ & $37.7$\\
    \bottomrule
\end{tabular}
\label{tab:comparison_attention}
\end{table}

\textbf{Comparisons of Query Features.}
\begin{figure}
    \centering
    \includegraphics[width=0.85\linewidth]{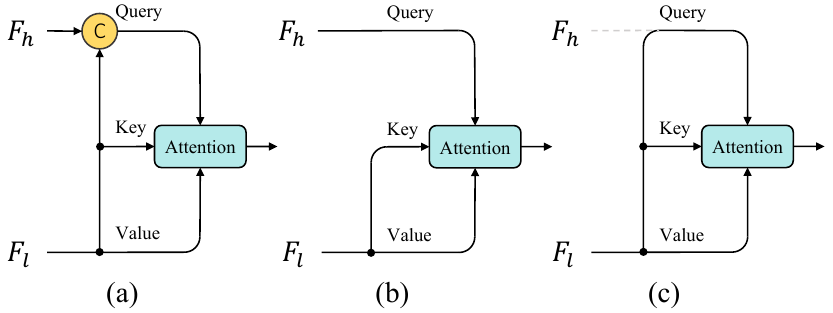}
    \caption{\textbf{Query features.} (a) query features is the concatenation of high-resolution features and low-resolution features. (b) query features consists only high-resolution features. (c) query features consists only low-resolution features.}
    \label{fig:gai_analysis}
\end{figure}
To further investigate the Guided Attentive Interpolation, we evaluate the performance of the attention module with different \textit{query} features. 
In the proposed Guided Attentive Interpolation, the combination of high-resolution feature maps and low-resolution feature maps acts as the \textit{query} features. Fig.~\ref{fig:gai_analysis} illustrates another two variants (b,c) of Guided Attentive Interpolation, which both use features of a single resolution.
When using only $F_l$ feature maps as \textit{query}, illustrated in Fig.~\ref{fig:gai_analysis}(c), the Guided Attentive Interpolation can be viewed as a self-attention on the low-resolution feature maps, which only handle the dependencies and aggregate features in the same level without the guidance from the high-resolution feature maps. However, using only $F_h$ feature maps, shown in Fig.~\ref{fig:gai_analysis}(b), will lose the gain from self-attention.

Tab.~\ref{tab:comparison_query} shows the segmentation results using different \textit{query} features and the results can validate the effectiveness of using the combination of low-resolution and high-resolution as the \textit{query} features. High-resolution features can provide spatial relations and low-resolution features will bring more semantic relations. The combination of semantic and spatial relations will contribute to better representations for each pixel, thus making it feasible for building fine-grained semantic features for dense prediction tasks.
\begin{table}
\caption{\textbf{Query Features.} Comparison of different query features (\{a,b,c\}).}
\centering
\small
\renewcommand\arraystretch{1.2}
\setlength{\tabcolsep}{3mm}
\begin{tabular}{c|ccc}
    \toprule
    \textit{query} & $F_l$ & $F_h$ & $(F_h, F_l)$ \\
    \hline
    
    \hline
    mIoU(\%) & $77.9$ & $77.7$ & $78.8$ \\
    \bottomrule
\end{tabular}
\label{tab:comparison_query}
\end{table}
\section{Conclusion}
We propose Guided Attentive Interpolation to enhance the feature representation by aggregating low-resolution features according to the pairwise relations of high-resolution pixels. 
It is a novel and effective replacement for traditional coordinate-based feature upsampling/aggregation operations. 
The Guided Attentive Interpolation can simultaneously interpolate the low-resolution features to high-resolution feature maps and enrich more semantics, making it feasible to obtain fine-grained semantic features. 
Extensive experiments on the standard driving scene parsing benchmark show that Guided Attentive Interpolation makes the simple ResNet-18 or DF-2 achieve accurate and fast segmentation results that are on-par with previous state-of-the-art methods. 
As a plug-in high-resolution feature reassembly operation, we believe Guided Attentive Interpolation can be widely applied in dense/structural prediction deep networks.

\backmatter
\bmhead{Acknowledgements}
The authors would like to thank Zilong Huang for the helpful discussion.
\bibliographystyle{unsrt}
\bibliography{sn-bibliography}

\end{document}